\documentclass[conference]{IEEEtran}
\IEEEoverridecommandlockouts
\usepackage{booktabs}
\usepackage{tabularx}
\usepackage{cite}
\usepackage{amsmath,amssymb,amsfonts}
\usepackage{algorithmic}
\usepackage[ruled,vlined]{algorithm2e}
\usepackage{graphicx}
\usepackage{textcomp}
\usepackage{xcolor}
\usepackage{pifont}
\usepackage{enumitem}
\usepackage{bm}
\usepackage{hyperref}

\def\BibTeX{{\rm B\kern-.05em{\sc i\kern-.025em b}\kern-.08em
    T\kern-.1667em\lower.7ex\hbox{E}\kern-.125emX}}
\begin{document}
\title{Mamba-YOLO-World: Marrying YOLO-World with Mamba for Open-Vocabulary Detection}

\author{
\IEEEauthorblockN{
        Haoxuan Wang$^{1}$,
        Qingdong He$^{2}$,
        Jinlong Peng$^{2}$,
        Hao Yang$^{3}$,
        Mingmin Chi$^{1,4 *}$ \thanks{\textsuperscript{$*$}Corresponding author (mmchi@fudan.edu.cn).},
        Yabiao Wang$^{2}$
    }
\\
\IEEEauthorblockA{
   \textit{$^{1}$School of computer science, Shanghai key laboratory of data science, Fudan University, China} \\
    \textit{$^{2}$Tencent Youtu Lab, China} \quad
    \textit{$^{3}$Shanghai Jiao Tong University, China} \\
    \textit{$^{4}$Zhongshan PoolNet Technology Co., Ltd, Zhongshan Fudan Joint Innovation Center, China }
    }
}
\maketitle

\begin{abstract}
Open-vocabulary detection (OVD) aims to detect objects beyond a predefined set of categories. As a pioneering model incorporating the YOLO series into OVD, YOLO-World is well-suited for scenarios prioritizing speed and efficiency. However, its performance is hindered by its neck feature fusion mechanism, which causes the quadratic complexity and the limited guided receptive fields. To address these limitations, we present Mamba-YOLO-World, a novel YOLO-based OVD model employing the proposed MambaFusion Path Aggregation Network (MambaFusion-PAN) as its neck architecture. Specifically, we introduce an innovative State Space Model-based feature fusion mechanism consisting of a Parallel-Guided Selective Scan algorithm and a Serial-Guided Selective Scan algorithm with linear complexity and globally guided receptive fields. It leverages multi-modal input sequences and mamba hidden states to guide the selective scanning process. Experiments demonstrate that our model outperforms the original YOLO-World on the COCO and LVIS benchmarks in both zero-shot and fine-tuning settings while maintaining comparable parameters and FLOPs. Additionally, it surpasses existing state-of-the-art OVD methods with fewer parameters and FLOPs.
\end{abstract}


\renewcommand{\thefootnote}{}
\footnotetext{Code is available at: https://github.com/Xuan-World/Mamba-YOLO-World.}
\renewcommand{\thefootnote}{\arabic{footnote}}

\section{Introduction}

Object detection, as a fundamental task in computer vision, plays a crucial role in various domains such as autonomous vehicles, personal electronic devices, healthcare, and security. The traditional methods\cite{fast-rcnn, faster-rcnn, yolo,detr, deformable-detr, dino} have
made great progress in object detection. Nevertheless, these models are trained on closed-set datasets, limiting their capabilities to predefined categories (e.g., 80 categories in the COCO\cite{coco} dataset). To overcome such limitations, open-vocabulary detection (OVD)\cite{OVR-CNN} has emerged as a new task that requires the model to detect objects beyond a predefined set of categories.

\begin{figure}[ht]
    \centering
    \includegraphics[width=1\linewidth]{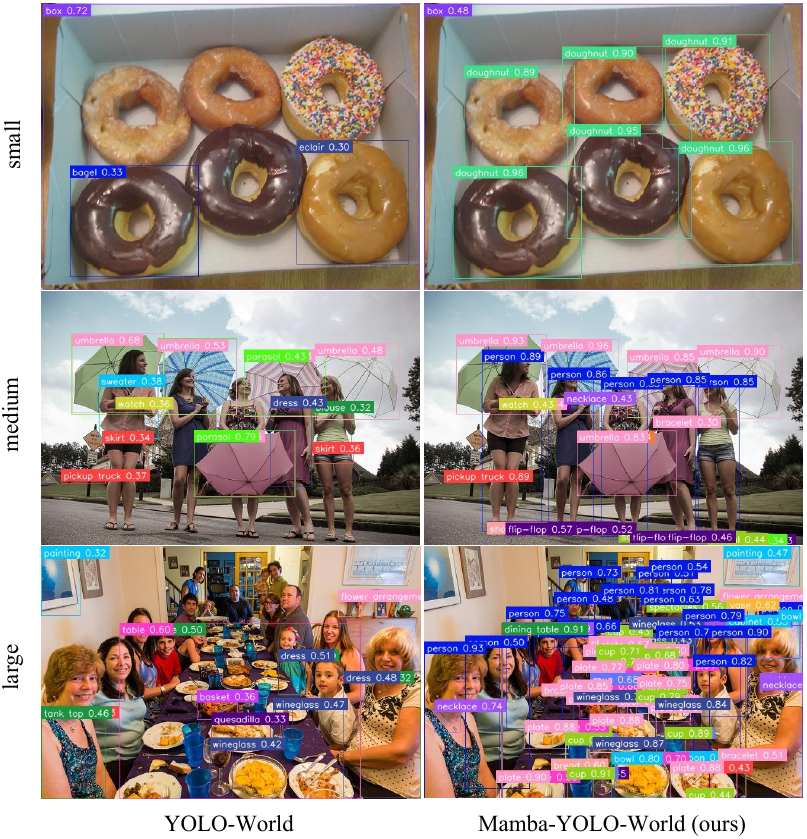}
    \vspace{-20pt}
    \caption{Visualization Results of Zero-shot Inference on LVIS\cite{lvis}. Our Mamba-YOLO-World significantly outperforms YOLO-World in terms of accuracy and generalization across small, medium, and large models.}
    \vspace{-10px}
    \label{fig:visualization}
\end{figure}

Some previous OVD works\cite{vild, baron, fvlm, dstdet,cora} attempt to leverage the inherent image-text alignment capabilities of pre-trained Vision-Language Models (VLMs). However, these VLMs are trained primarily at the image-text level, resulting in a lack of alignment capabilities at the region-text level. Recent works, such as MDETR\cite{mdetr}, GLIP\cite{glip}, DetClip\cite{detclip}, Grounding DINO\cite{grounding-dino},  mm-Grounding-DINO\cite{mmGroundingDINO} and YOLO-World\cite{yoloworld} redefine OVD as a vision-language pre-training task, employing traditional object detectors to directly learn region-text level open-vocabulary alignment capability on large-scale datasets.

According to the aforementioned related works, the key to converting a traditional object detector into an OVD model lies in implementing a visual-linguistic feature fusion mechanism that is adaptable to the existing neck structure of the model, such as the VL-PAN\cite{yoloworld} in YOLO-World and the Feature-Enhancer\cite{grounding-dino} in Grounding-DINO. As a pioneering model incorporating the YOLO series into OVD, YOLO-World is well-suited for deployment in scenarios prioritizing speed and efficiency. Despite this, its performance is hindered by its VL-PAN feature fusion mechanism.

Specifically, the VL-PAN employs a max-sigmoid visual channel attention mechanism in text-to-image feature fusion flow and a multi-head cross-attention mechanism in image-to-text fusion flow, leading to several limitations. 
\textit{Firstly}, the complexities of both fusion flows increase quadratically with the product of image size and text length, due to the cross-modal attention mechanism.
\textit{Secondly}, the VL-PAN lacks globally guided receptive fields. On the one hand, the text-to-image fusion flow solely generates a visual channel weighting vector that lacks spatial guidance at the pixel level.
On the other hand, the image-to-text fusion flow merely allows image information to guide each word individually, failing to leverage the contextual information within text descriptions.

To address the above limitations, we introduce Mamba-YOLO-World, a novel YOLO-based OVD model employing the proposed MambaFusion Path Aggregation Network (MambaFusion-PAN) as its neck architecture. Recently, Mamba\cite{mamba}, as an emerging State Space Model (SSM), has demonstrated its ability to avoid quadratic complexity and capture global receptive fields\cite{vmamba,vim,mambavision,samba,mamba2}. However, simply concatenating the multi-modal features in Mamba\cite{VLMamba,meteor,cobra}  results in a complexity of 
 $O(N+M)$, which increases proportionally with the length of the concatenated sequence. This is particularly problematic for large vocabulary in OVD. Motivated by it, we propose a State Space Model-based feature fusion mechanism in MambaFusion-PAN. We use the mamba hidden state as an intermediary for feature fusion between different modalities, which incurs $O(N+1)$ complexity and provides globally guided receptive fields. 
 The visualization results shown in Fig. \ref{fig:visualization} demonstrate that our Mamba-YOLO-World significantly outperforms YOLO-World in terms of accuracy and generalization across all size variants.

Our contributions can be summarized as follows:
\begin{itemize}[left=0pt]
    \item We present Mamba-YOLO-World, a novel YOLO-based OVD model employing the proposed MambaFusion-PAN as its neck architecture. 
    \item We introduce a State Space Model-based feature fusion mechanism consisting of a Parallel-Guided Selective Scan algorithm and a Serial-Guided Selective Scan algorithm,  with $O(N+1)$ complexity and globally guided receptive fields.
    \item Experiments demonstrate that our model outperforms the original YOLO-World while maintaining comparable parameters and FLOPs. Additionally, it surpasses existing state-of-the-art OVD methods with fewer parameters and FLOPs.
\end{itemize}

\begin{figure*}[t]
  \centering
  \includegraphics[width=\textwidth]{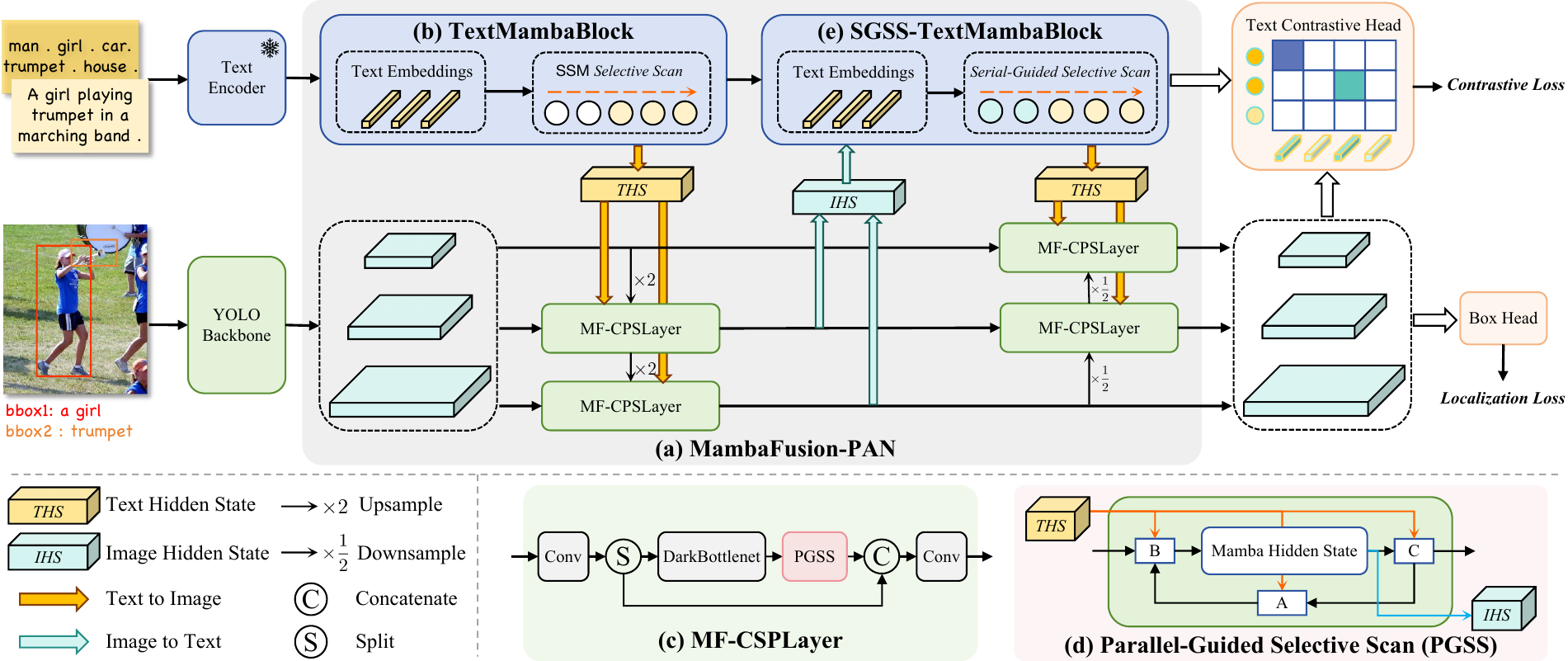}
    \vspace{-15pt}
  \caption{
      Overall Architecture of Mamba-YOLO-World. It consists of five key components: 
      \textbf{(a)} MambaFusion-PAN is our proposed feature fusion network for replacing the Path Aggregation Feature Pyramid Network in YOLO.
      \textbf{(b)} TextMambaBlock comprises stacked Mamba layers scanning the input text embeddings to extract the output text features and text hidden state (\textit{THS}). 
      \textbf{(c)} MF-CSPLayer incorporates the proposed PGSS algorithm into a YOLO CSPLayer style network. 
      \textbf{(d)} In the Parallel-Guided Selective Scan (PGSS) algorithm, the compressed textual information \textit{THS} is injected into Mamba parameters in \textit{parallel} with the entire visual selective scanning process to extract the output image features and image hidden state (\textit{IHS}). 
      \textbf{(e)} SGSS-TextMambaBlock is a TextMambaBlock with a Serial-Guided Selective Scan algorithm. It adjusts Mamba parameters in \textit{serial} by scanning the compressed visual information \textit{IHS}  before extracting the text features.
  }
    \vspace{-10pt}
  \label{fig:overall}
\end{figure*}

\section{Method}

Mamba-YOLO-World is mainly developed based on YOLOv8\cite{yolov8_ultralytics}, comprising a Darknet Backbone\cite{yolo} and a CLIP\cite{clip} Text Encoder as model's backbone, our MambaFusion-PAN as model's neck, and a text contrastive classification head along with a bounding box regression head as model's heads, as depicted in Fig. \ref{fig:overall}. 

\subsection {Mamba Preliminaries}
 For a continuous input signal $ u(t) \in \mathbb{R} $, SSM\cite{s4} maps it to a continuous output signal $y(t) \in \mathbb{R}$  through a hidden state $h(t) \in \mathbb{R}^E $.
 \vspace{-5px}
\begin{align}
    h^{'}(t) &= A h(t) + B u(t) \\
    y(t) &= C h(t)
\end{align}
where $E$ is the SSM state expansion factor, $A \in \mathbb{R}^{E\times E}$  is the state transition matrix, and \( B \in \mathbb{R}^{E \times 1} \) and \( C \in \mathbb{R}^{1 \times E} \) are the input and output mapping matrices, respectively. 
Building on SSM, Mamba\cite{mamba} introduces the Selective Scan algorithm, making \( A \), \( B \), and \( C \) functions of the input sequence.

\subsection{MambaFusion-PAN}
The MambaFusion-PAN is our proposed feature fusion network for replacing the Path Aggregation Feature Pyramid Network in YOLO. As shown in Fig. \ref{fig:overall}(a), the MambaFusion-PAN utilizes the proposed SSM-based \textit{parallel} and \textit{serial} feature fusion mechanism to aggregate multi-scale image features and enhance text features simultaneously through a three-stage feature fusion flow between visual and linguistic branch: Text-to-Image, Image-to-Text, and finally Text-to-Image. Specific components are detailed in the following parts of this section.

\subsubsection{Mamba Hidden State}
Currently, both Transformer-based and Mamba-based VLMs simply concatenate multi-modal features \cite{grounding-dino, mmGroundingDINO, LLaVa, blip, VLMamba, meteor, cobra}, leading to an inevitable increase in complexity as the text sequence length and image resolution grow. Although VL-PAN in YOLO-World employs unidirectional fusion without feature concatenation, it still results in $O(N^2)$ complexity. This is due to the visual channel attention mechanism in the text-to-image fusion flow and the multi-head cross-attention mechanism in the image-to-text fusion flow.

To address these issues, we propose extracting the compressed sequence information through the mamba hidden state $h(t)\in \mathbb{R}^{D \times E}$ to serve as an intermediary for feature fusion between different modalities, where $D$ is the dimension of the input sequence and $E$ is the SSM state expansion factor\cite{mamba,mamba2}. Since both $D$ and $E$ are constants and not affected by the length of the sequences, our feature fusion mechanism has a complexity of \(O(N+1)\), where $N$ comes from the input sequence of one modality and $1$ comes from the mamba hidden state of another modality.

\subsubsection {TextMambaBlock}
The TextMambaBlock is composed of stacked Mamba layers.
Given the text embeddings \( w_0 \mathbin{\in} \mathbb{R}^{L_t\times D_t} \) output from the CLIP Text Encoder, we adopt the TextMambaBlock depicted in Fig. \ref{fig:overall}(b) to extract not only the output text features \( w_1\mathbin{\in} \mathbb{R}^{L_t\times D_t} \) but also the text hidden state $\textit{THS}\mathbin{\in} \mathbb{R}^{D_t\times E_t}$, which will be used for subsequent Text-to-Image feature fusion.

\subsubsection {MF-CSPLayer}
As shown in Fig. \ref{fig:overall}(c), we integrate \textit{THS} with multi-scale image features through the MambaFusion CSPLayer (MF-CSPLayer). The MF-CSPLayer incorporates the proposed Parallel-Guided Selective Scan algorithm into a YOLO CSPLayer style network. After processing through MF-CSPLayer, we can obtain not only the output image features but also the image hidden state $\textit{IHS} \mathbin{\in} \mathbb{R}^{D_i \times E_i}$, which will be used for subsequent Image-to-Text feature fusion.

\subsubsection {Parallel-Guided Selective Scan}
 The Mamba Selective Scan algorithm dynamically adjusts internal parameters based on the input sequence. Motivated by this, we innovatively propose the Parallel-Guided Selective Scan (PGSS) algorithm, which dynamically adjusts the values of Mamba internal parameters (A, B, and C) based on both the input image sequence and \textit{THS} during the scanning process, as illustrated in Fig. \ref{fig:overall}(d) and Algorithm \ref{alg:gss}. 
 Therefore, the compressed textual information is injected into Mamba in \textit{parallel} with the entire visual selective scanning process, enabling the multi-scale image features to be guided at the pixel level rather than the channel level. 
The outputs generated from it are passed to the subsequent layers of MF-CSPLayer. 
In the following, we refer to this part as the Text-to-Image feature fusion flow.

\vspace{-10pt}
\begin{algorithm} [h]
    \small
    \SetAlgoLined
    \label{alg:gss}
    \caption{Parallel-Guided Selective Scan}
    \KwIn{$\mathbf{X} \in \mathbb{R}^{L_i \times D_i}, \bm{\mathit{THS}} \in \mathbb{R}^{D_t\times E_t}$}
    \KwOut{$\mathbf{Y} \in \mathbb{R}^{L_i \times D_i},\bm{\mathit{IHS}} \in \mathbb{R}^{D_i\times E_i}$}
    $\mathbf{A}: size(D,E) \gets \mathrm{Parameter} $ \\
    $\mathbf{B}: size(L,E) \gets \mathrm{Linear_B(\mathbf{X}, \bm{\mathit{THS}})}$ \\
    $\mathbf{C}: size(L,E) \gets \mathrm{Linear_C(\mathbf{X}, \bm{\mathit{THS}})}$ \\
    $\mathbf{\Delta}: size(L,D) \gets \mathrm{softplus (Linear_\Delta(\mathbf{X},\bm{\mathit{THS}})\!+\!Param_\Delta)}$ \\
    $\mathbf{\overline{A},\overline{B}}: size(L,D,E) \gets \mathrm{discretize(\mathbf{\Delta,A,B})}$ \\
    $\mathbf{Y},\bm{\mathit{IHS}} \gets \mathrm{SSM(\mathbf{\overline{A},\overline{B},C})(\mathbf{X})}$\\
    \textbf{return}$\quad \mathbf{Y},\bm{\mathit{IHS}}$
\end{algorithm}
\vspace{-10pt}

\subsubsection {Serial-Guided Selective Scan}
\label{ssec:subhead}
The Mamba Selective Scan algorithm continuously compresses information into \( h(t) \) based on the input sequence. Motivated by this, we propose the Serial-Guided Selective Scan (SGSS) algorithm and combine it into the TextMambaBlock, as represented in Fig. \ref{fig:overall}(e). 
The SGSS aims to compress the prior knowledge from preceding sequences into \(h(t)\) and use it as a guidance for the following sequences. 
Specifically, the SGSS-TextMambaBlock adjusts the values of Mamba internal parameters (A, B, and C) in \textit{serial} by scanning the compressed visual information \textit{IHS}  before extracting the text features.
In the following, we refer to this part as the Image-to-Text feature fusion flow.

\section{Experiment}
\subsection {Implementation Details}
Mamba-YOLO-World is developed based on the MMYOLO\cite{mmyolo} toolbox and the MMDetection\cite{mmdetection} toolbox. We provide three size variants, i.e., small (S), medium (M), and large (L). The experiments involve a pre-training stage followed by a fine-tuning stage. During the pre-training stage, we adopt the detection and grounding datasets including Objects365 (V1)\cite{objects365}, GQA\cite{gqa}, and Flickr30k\cite{flickr30k}. In line with other OVD methods\cite{yoloworld,mdetr,grounding-dino,mmGroundingDINO,glip,detclip}, the GQA and Flickr30k datasets are collectively designated as the GoldG\cite{mdetr} dataset after excluding images from COCO\cite{coco}. During the fine-tuning stage, we use the pre-trained Mamba-YOLO-World and fine-tune it on the downstream task datasets. Unless specified, we conduct the experiments following the settings of YOLO-World\cite{yoloworld}.

\begin{table*}[t] 
\centering
\caption{Zero-shot Evaluation on LVIS minival (\%)}
\vspace{-5pt}
\label{tab:zs-lvis}
\resizebox{\textwidth}{!}{
    \begin{tabularx}{\textwidth}{ p{4cm} p{2.5cm} p{1.5cm} p{1.5cm} X p{0.7cm}p{0.7cm}p{0.7cm}p{0.7cm}}
    \toprule
    Method & Backbone & Params & FLOPs & Pre-trained Data & $AP$ & $AP_r$ & $AP_c$ & $AP_f$  \\
    \midrule
    MDETR\cite{mdetr}  & R-101\cite{resnet} & 169M & - & GoldG & 16.7 & 11.2 & 14.6 & 19.5 \\
    GLIP-T\cite{glip} & Swin-T\cite{swin}  & 232M & - & O365,GoldG & 24.9 & 17.7 & 19.5 & 31.0 \\
    Grounding-DINO-T\cite{grounding-dino} & Swin-T\cite{swin} & 172M & - & O365,GoldG & 25.6 & 14.4 & 19.6 & 32.2 \\
    DetCLIP-T\cite{detclip} & Swin-T\cite{swin} & 155M & - & O365,GoldG & 34.4 & 26.9 & 33.9 & 36.3 \\
    mm-Grounding-DINO-T\cite{mmGroundingDINO} & Swin-T\cite{swin} & 173M &  -  & O365,GoldG & 35.7 & 28.1 & 30.2 & 42.0 \\
    \midrule
    YOLO-World-S\cite{yoloworld} & YOLOv8-S\cite{yolov8_ultralytics} & 77M & 297G & O365,GoldG & 26.2 & 19.1 & 23.6 & 29.8	 \\
    Mamba-YOLO-World-S (ours)  & YOLOv8-S\cite{yolov8_ultralytics} & 78M & 297G &  O365,GoldG & \textbf{27.7} & \textbf{19.5} & \textbf{27.0} & \textbf{29.9} \\
    \midrule
    YOLO-World-M\cite{yoloworld} & YOLOv8-M\cite{yolov8_ultralytics} & 92M & 324G &  O365,GoldG & 31.0 & 23.8 & 29.2 & 33.9\\
    Mamba-YOLO-World-M (ours)  & YOLOv8-M\cite{yolov8_ultralytics} & 94M & 324G &  O365,GoldG & \textbf{32.8} & \textbf{27.0} & \textbf{31.9} & \textbf{34.8} \\
    \midrule
    YOLO-World-L\cite{yoloworld} & YOLOv8-L\cite{yolov8_ultralytics} & 111M & 370G &  O365,GoldG & 35.0 & 27.1 & 32.8 & \textbf{38.3} \\
    Mamba-YOLO-World-L (ours)  & YOLOv8-L\cite{yolov8_ultralytics} & 113M & 369G &  O365,GoldG & \textbf{35.0} & \textbf{29.3} & \textbf{34.2} &  36.8 \\
    \bottomrule
    \end{tabularx}
}
    \vspace{-15pt}
\end{table*}

\subsection{Zero-shot Results}
After pre-training, we directly evaluate the proposed Mamba-YOLO-World on both LVIS\cite{lvis} and COCO\cite{coco} benchmarks in a zero-shot manner and provide a comprehensive comparison with YOLO-World and other existing state-of-the-art methods.

\subsubsection{Zero-shot Evaluation on LVIS}
 The LVIS dataset encompasses 1203 long-tail object categories. Following previous works\cite{yoloworld,mdetr,grounding-dino,mmGroundingDINO,detclip,glip}, we use the \textit{Fixed AP}\cite{fixedAP} metric and report 1000 predictions per image on LVIS \texttt{minival} for a fair comparison. According to Table \ref{tab:zs-lvis}, Mamba-YOLO-World achieves a \(+1.5\%\) AP improvement for small variant and a \(+1.8\%\) AP improvement for medium variant compared to YOLO-World while keeping comparable parameters and FLOPs. Moreover, it outperforms YOLO-World by \(+0.4\% \!\sim\! +3.2\%\) AP\(_r\) and \(+1.4\% \!\sim\! +3.4\%\) AP\(_c\) across all size variants. The Mamba-YOLO-World-L obtains superior results compared with previous state-of-the-art methods such as \cite{mdetr,glip,grounding-dino,detclip} with fewer parameters and FLOPs.
 
\subsubsection{Zero-shot Evaluation on COCO}
The COCO dataset contains 80 categories and is the most commonly used dataset for object detection. As illustrated in Table \ref{tab:zs-coco}, our Mamba-YOLO-World shows overall advantages, outperforming YOLO-World by \(+0.4\%\!\sim\!+1\%\) AP across all size variants.

\begin{table}[t]
\centering
\caption{Zero-shot Evaluation on COCO (\%)}
    \vspace{-5pt}
    \label{tab:zs-coco} 
    \begin{tabularx}{\columnwidth}{ p{4cm} p{1.5cm} XXX}
    \toprule
    Method & Pre-train & $AP$ & $AP_{50}$ & $AP_{75}$  \\
    \midrule
    YOLO-World-S\cite{yoloworld} & O365,GoldG & 37.6 & 52.3 & 40.7 \\
    Mamba-YOLO-World-S (ours) & O365,GoldG &  \textbf{38.0} & \textbf{52.9} & \textbf{41.0} \\
    \midrule
    YOLO-World-M\cite{yoloworld} & O365,GoldG & 42.8 & 58.3 & 46.4\\
    Mamba-YOLO-World-M (ours)  &O365,GoldG & \textbf{43.2} & \textbf{58.8} & \textbf{46.6}  \\
    \midrule
    YOLO-World-L\cite{yoloworld}  & O365,GoldG &  44.4 & 59.8 & 48.3 \\
    Mamba-YOLO-World-L (ours)  & O365,GoldG & \textbf{45.4} & \textbf{61.3} & \textbf{49.4} \\
    \bottomrule
    \end{tabularx}
\vspace{-5pt}
\end{table}

\subsection{Fine-tuning Results}
In Table \ref{tab:ft-coco}, we further evaluate the fine-tuning results on the COCO benchmark. After fine-tuning on COCO \texttt{train2017}, Mamba-YOLO-World achieves higher accuracy and consistently outperforms the fine-tuned YOLO-World by \(+0.2\%\!\sim\!+0.8\%\) AP across all size variants.

\begin{table}[t]
\centering
\caption{Fine-tuning Evaluation on COCO (\%)}
    \vspace{-5pt}
    \label{tab:ft-coco}
    \begin{tabularx}{\columnwidth}{ p{4cm} p{1.5cm}XXX}
    \toprule
    Method & Pre-train & $AP$ & $AP_{50}$ & $AP_{75}$  \\
    \midrule
    YOLO-World-S\cite{yoloworld}& O365,GoldG & 45.9 & 62.3 & 50.1	 \\
    Mamba-YOLO-World-S (ours)&O365,GoldG & \textbf{46.4} & \textbf{62.5} & \textbf{50.5} \\
    \midrule
    YOLO-World-M\cite{yoloworld}& O365,GoldG & 51.2 & 68.1 & 55.9 \\
    Mamba-YOLO-World-M (ours) &O365,GoldG & \textbf{51.4} & \textbf{68.2} &  \textbf{56.1}  \\
    \midrule
    YOLO-World-L\cite{yoloworld}& O365,GoldG & 53.3 & 70.1 & 58.2 \\
    Mamba-YOLO-World-L (ours)& O365,GoldG & \textbf{54.1} & \textbf{71.1} & \textbf{59.0} \\
    \bottomrule
    \end{tabularx}
\vspace{-10pt}
\end{table}

\subsection{Ablation Studies}
 In Table \ref{tab:ablation}, we conduct ablation experiments to analyze the impact of the MambaFusion-PAN Text-to-Image and Image-to-Text feature fusion flow based on Mamba-YOLO-World-S. The zero-shot evaluation results on the COCO benchmark indicate that both our parallel (Text$\rightarrow$Image) and serial (Image$\rightarrow$Text) feature fusion methods effectively boost the performance without increasing the parameters or FLOPs.

\begin{table}[t]
    \centering
    \caption{Ablation on MambaFusion-PAN}
    \vspace{-4pt}
    \begin{tabularx}{\columnwidth}{cc p{0.7cm} p{0.7cm} XXX}
        \toprule
        Text$\rightarrow$Image & Image$\rightarrow$Text & Params & FLOPs & $AP$ & $AP_{50}$ & $AP_{75}$ \\
        \midrule
        \ding{55} & \ding{55} & 78M & 297G & 37.1 & 51.7 & 39.9 \\
        \ding{55} & \ding{51} & 78M & 297G & 37.3 & 52.0 & 40.2 \\
        \ding{51} & \ding{55} & 78M & 297G & 37.8 & 52.5 & 40.7 \\
        \ding{51} & \ding{51} & 78M & 297G & \textbf{38.0} & \textbf{52.9} & \textbf{41.0} \\
        \bottomrule
    \end{tabularx}
    \label{tab:ablation}
\end{table}

\begin{figure}[t]
    \centering     
    \includegraphics[width=0.9\linewidth]{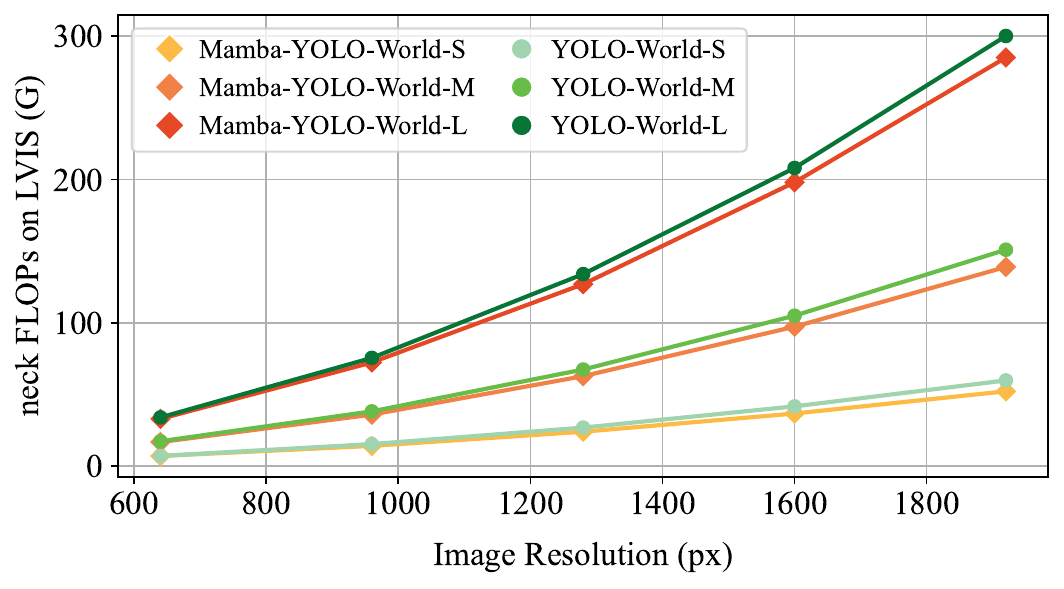}
    \vspace{-10px}
    \caption{Comparison of Neck FLOPs Across Different Image Resolutions}
    \vspace{-5px}
    \label{fig:flops}
\end{figure}

 Additionally, we analyze the changes in computational cost as the input image resolution increases. As illustrated in Fig. \ref{fig:flops}, the MambaFusion-PAN (neck of Mamba-YOLO-World) consumes up to \(15\%\) fewer FLOPs than the VL-PAN (neck of YOLO-World) across all size variants, indicating a lower model complexity of our MambaFusion-PAN.
 
\section{Conclusion}
In this paper, we present Mamba-YOLO-World for open-vocabulary object detection. We introduce an innovative State Space Model-based feature fusion mechanism and integrate it into MambaFusion-PAN. Experimental results demonstrate that Mamba-YOLO-World outperforms the original YOLO-World with comparable parameters and FLOPs. We hope this work will bring new insights into the multi-modal Mamba architecture and encourage further exploration for open-vocabulary vision tasks.

\vfill\pagebreak
\bibliographystyle{IEEEtran}
\bibliography{refs}

\begin{thebibliography}{10}
\providecommand{\url}[1]{#1}
\csname url@samestyle\endcsname
\providecommand{\newblock}{\relax}
\providecommand{\bibinfo}[2]{#2}
\providecommand{\BIBentrySTDinterwordspacing}{\spaceskip=0pt\relax}
\providecommand{\BIBentryALTinterwordstretchfactor}{4}
\providecommand{\BIBentryALTinterwordspacing}{\spaceskip=\fontdimen2\font plus
\BIBentryALTinterwordstretchfactor\fontdimen3\font minus \fontdimen4\font\relax}
\providecommand{\BIBforeignlanguage}[2]{{%
\expandafter\ifx\csname l@#1\endcsname\relax
\typeout{** WARNING: IEEEtran.bst: No hyphenation pattern has been}%
\typeout{** loaded for the language `#1'. Using the pattern for}%
\typeout{** the default language instead.}%
\else
\language=\csname l@#1\endcsname
\fi
#2}}
\providecommand{\BIBdecl}{\relax}
\BIBdecl

\bibitem{fast-rcnn}
R.~Girshick, ``Fast r-cnn,'' in \emph{Proceedings of the IEEE international conference on computer vision}, 2015, pp. 1440--1448.

\bibitem{faster-rcnn}
S.~Ren, K.~He, R.~Girshick, and J.~Sun, ``Faster r-cnn: Towards real-time object detection with region proposal networks,'' \emph{Advances in neural information processing systems}, vol.~28, 2015.

\bibitem{yolo}
J.~Redmon, S.~Divvala, R.~Girshick, and A.~Farhadi, ``You only look once: Unified, real-time object detection,'' in \emph{Proceedings of the IEEE conference on computer vision and pattern recognition}, 2016, pp. 779--788.

\bibitem{detr}
N.~Carion, F.~Massa, G.~Synnaeve, N.~Usunier, A.~Kirillov, and S.~Zagoruyko, ``End-to-end object detection with transformers,'' in \emph{European conference on computer vision}.\hskip 1em plus 0.5em minus 0.4em\relax Springer, 2020, pp. 213--229.

\bibitem{deformable-detr}
X.~Zhu, W.~Su, L.~Lu, B.~Li, X.~Wang, and J.~Dai, ``Deformable {DETR}: Deformable transformers for end-to-end object detection,'' in \emph{International Conference on Learning Representations}, 2021.

\bibitem{dino}
H.~Zhang, F.~Li, S.~Liu, L.~Zhang, H.~Su, J.~Zhu, L.~Ni, and H.-Y. Shum, ``{DINO}: {DETR} with improved denoising anchor boxes for end-to-end object detection,'' in \emph{The Eleventh International Conference on Learning Representations}, 2023.

\bibitem{coco}
T.-Y. Lin, M.~Maire, S.~Belongie, J.~Hays, P.~Perona, D.~Ramanan, P.~Doll{\'a}r, and C.~L. Zitnick, ``Microsoft coco: Common objects in context,'' in \emph{Computer Vision--ECCV 2014: 13th European Conference, Zurich, Switzerland, September 6-12, 2014, Proceedings, Part V 13}.\hskip 1em plus 0.5em minus 0.4em\relax Springer, 2014, pp. 740--755.

\bibitem{OVR-CNN}
A.~Zareian, K.~D. Rosa, D.~H. Hu, and S.-F. Chang, ``Open-vocabulary object detection using captions,'' in \emph{Proceedings of the IEEE/CVF Conference on Computer Vision and Pattern Recognition}, 2021, pp. 14\,393--14\,402.

\bibitem{lvis}
A.~Gupta, P.~Dollar, and R.~Girshick, ``Lvis: A dataset for large vocabulary instance segmentation,'' in \emph{Proceedings of the IEEE/CVF conference on computer vision and pattern recognition}, 2019, pp. 5356--5364.

\bibitem{vild}
X.~Gu, T.-Y. Lin, W.~Kuo, and Y.~Cui, ``Open-vocabulary object detection via vision and language knowledge distillation,'' in \emph{International Conference on Learning Representations}, 2022.

\bibitem{baron}
S.~Wu, W.~Zhang, S.~Jin, W.~Liu, and C.~C. Loy, ``Aligning bag of regions for open-vocabulary object detection,'' in \emph{Proceedings of the IEEE/CVF conference on computer vision and pattern recognition}, 2023, pp. 15\,254--15\,264.

\bibitem{fvlm}
W.~Kuo, Y.~Cui, X.~Gu, A.~Piergiovanni, and A.~Angelova, ``Open-vocabulary object detection upon frozen vision and language models,'' in \emph{The Eleventh International Conference on Learning Representations}, 2023.

\bibitem{dstdet}
S.~Xu, X.~Li, S.~Wu, W.~Zhang, Y.~Li, G.~Cheng, Y.~Tong, K.~Chen, and C.~C. Loy, ``Dst-det: Simple dynamic self-training for open-vocabulary object detection,'' \emph{arXiv preprint arXiv:2310.01393}, 2023.

\bibitem{cora}
X.~Wu, F.~Zhu, R.~Zhao, and H.~Li, ``Cora: Adapting clip for open-vocabulary detection with region prompting and anchor pre-matching,'' in \emph{2023 IEEE/CVF Conference on Computer Vision and Pattern Recognition (CVPR)}, 2023, pp. 7031--7040.

\bibitem{mdetr}
A.~Kamath, M.~Singh, Y.~LeCun, G.~Synnaeve, I.~Misra, and N.~Carion, ``Mdetr-modulated detection for end-to-end multi-modal understanding,'' in \emph{Proceedings of the IEEE/CVF international conference on computer vision}, 2021, pp. 1780--1790.

\bibitem{glip}
L.~H. Li, P.~Zhang, H.~Zhang, J.~Yang, C.~Li, Y.~Zhong, L.~Wang, L.~Yuan, L.~Zhang, J.-N. Hwang \emph{et~al.}, ``Grounded language-image pre-training,'' in \emph{Proceedings of the IEEE/CVF Conference on Computer Vision and Pattern Recognition}, 2022, pp. 10\,965--10\,975.

\bibitem{detclip}
L.~Yao, J.~Han, Y.~Wen, X.~Liang, D.~Xu, W.~Zhang, Z.~Li, C.~Xu, and H.~Xu, ``Detclip: Dictionary-enriched visual-concept paralleled pre-training for open-world detection,'' \emph{Advances in Neural Information Processing Systems}, vol.~35, pp. 9125--9138, 2022.

\bibitem{grounding-dino}
S.~Liu, Z.~Zeng, T.~Ren, F.~Li, H.~Zhang, J.~Yang, C.~Li, J.~Yang, H.~Su, J.~Zhu \emph{et~al.}, ``Grounding dino: Marrying dino with grounded pre-training for open-set object detection,'' \emph{arXiv preprint arXiv:2303.05499}, 2023.

\bibitem{mmGroundingDINO}
X.~Zhao, Y.~Chen, S.~Xu, X.~Li, X.~Wang, Y.~Li, and H.~Huang, ``An open and comprehensive pipeline for unified object grounding and detection,'' \emph{arXiv preprint arXiv:2401.02361}, 2024.

\bibitem{yoloworld}
T.~Cheng, L.~Song, Y.~Ge, W.~Liu, X.~Wang, and Y.~Shan, ``Yolo-world: Real-time open-vocabulary object detection,'' in \emph{Proceedings of the IEEE/CVF Conference on Computer Vision and Pattern Recognition}, 2024, pp. 16\,901--16\,911.

\bibitem{mamba}
A.~Gu and T.~Dao, ``Mamba: Linear-time sequence modeling with selective state spaces,'' \emph{arXiv preprint arXiv:2312.00752}, 2023.

\bibitem{vmamba}
Y.~Liu, Y.~Tian, Y.~Zhao, H.~Yu, L.~Xie, Y.~Wang, Q.~Ye, and Y.~Liu, ``Vmamba: Visual state space model,'' \emph{arXiv preprint arXiv:2401.10166}, 2024.

\bibitem{vim}
L.~Zhu, B.~Liao, Q.~Zhang, X.~Wang, W.~Liu, and X.~Wang, ``Vision mamba: Efficient visual representation learning with bidirectional state space model,'' \emph{arXiv preprint arXiv:2401.09417}, 2024.

\bibitem{mambavision}
A.~Hatamizadeh and J.~Kautz, ``Mambavision: A hybrid mamba-transformer vision backbone,'' \emph{arXiv preprint arXiv:2407.08083}, 2024.

\bibitem{samba}
L.~Ren, Y.~Liu, Y.~Lu, Y.~Shen, C.~Liang, and W.~Chen, ``Samba: Simple hybrid state space models for efficient unlimited context language modeling,'' \emph{arXiv preprint arXiv:2406.07522}, 2024.

\bibitem{mamba2}
T.~Dao and A.~Gu, ``Transformers are ssms: Generalized models and efficient algorithms through structured state space duality,'' \emph{arXiv preprint arXiv:2405.21060}, 2024.

\bibitem{VLMamba}
Y.~Qiao, Z.~Yu, L.~Guo, S.~Chen, Z.~Zhao, M.~Sun, Q.~Wu, and J.~Liu, ``Vl-mamba: Exploring state space models for multimodal learning,'' \emph{arXiv preprint arXiv:2403.13600}, 2024.

\bibitem{meteor}
B.-K. Lee, C.~W. Kim, B.~Park, and Y.~M. Ro, ``Meteor: Mamba-based traversal of rationale for large language and vision models,'' \emph{arXiv preprint arXiv:2405.15574}, 2024.

\bibitem{cobra}
H.~Zhao, M.~Zhang, W.~Zhao, P.~Ding, S.~Huang, and D.~Wang, ``Cobra: Extending mamba to multi-modal large language model for efficient inference,'' \emph{arXiv preprint arXiv:2403.14520}, 2024.

\bibitem{yolov8_ultralytics}
\BIBentryALTinterwordspacing
G.~Jocher, A.~Chaurasia, and J.~Qiu, ``Ultralytics yolov8,'' 2023. [Online]. Available: \url{https://github.com/ultralytics/ultralytics}
\BIBentrySTDinterwordspacing

\bibitem{clip}
A.~Radford, J.~W. Kim, C.~Hallacy, A.~Ramesh, G.~Goh, S.~Agarwal, G.~Sastry, A.~Askell, P.~Mishkin, J.~Clark \emph{et~al.}, ``Learning transferable visual models from natural language supervision,'' in \emph{International conference on machine learning}.\hskip 1em plus 0.5em minus 0.4em\relax PMLR, 2021, pp. 8748--8763.

\bibitem{s4}
A.~Gu, K.~Goel, and C.~R\'e, ``Efficiently modeling long sequences with structured state spaces,'' in \emph{The International Conference on Learning Representations ({ICLR})}, 2022.

\bibitem{LLaVa}
H.~Liu, C.~Li, Q.~Wu, and Y.~J. Lee, ``Visual instruction tuning,'' \emph{Advances in neural information processing systems}, vol.~36, 2024.

\bibitem{blip}
J.~Li, D.~Li, C.~Xiong, and S.~Hoi, ``Blip: Bootstrapping language-image pre-training for unified vision-language understanding and generation,'' in \emph{International conference on machine learning}.\hskip 1em plus 0.5em minus 0.4em\relax PMLR, 2022, pp. 12\,888--12\,900.

\bibitem{mmyolo}
M.~Contributors, ``{MMYOLO: OpenMMLab YOLO} series toolbox and benchmark,'' \url{https://github.com/open-mmlab/mmyolo}, 2022.

\bibitem{mmdetection}
K.~Chen, J.~Wang, J.~Pang, Y.~Cao, Y.~Xiong, X.~Li, S.~Sun, W.~Feng, Z.~Liu, J.~Xu, Z.~Zhang, D.~Cheng, C.~Zhu, T.~Cheng, Q.~Zhao, B.~Li, X.~Lu, R.~Zhu, Y.~Wu, J.~Dai, J.~Wang, J.~Shi, W.~Ouyang, C.~C. Loy, and D.~Lin, ``{MMDetection}: Open mmlab detection toolbox and benchmark,'' \emph{arXiv preprint arXiv:1906.07155}, 2019.

\bibitem{objects365}
S.~Shao, Z.~Li, T.~Zhang, C.~Peng, G.~Yu, X.~Zhang, J.~Li, and J.~Sun, ``Objects365: A large-scale, high-quality dataset for object detection,'' in \emph{Proceedings of the IEEE/CVF international conference on computer vision}, 2019, pp. 8430--8439.

\bibitem{gqa}
D.~A. Hudson and C.~D. Manning, ``Gqa: A new dataset for real-world visual reasoning and compositional question answering,'' in \emph{Proceedings of the IEEE/CVF conference on computer vision and pattern recognition}, 2019, pp. 6700--6709.

\bibitem{flickr30k}
B.~A. Plummer, L.~Wang, C.~M. Cervantes, J.~C. Caicedo, J.~Hockenmaier, and S.~Lazebnik, ``Flickr30k entities: Collecting region-to-phrase correspondences for richer image-to-sentence models,'' in \emph{Proceedings of the IEEE international conference on computer vision}, 2015, pp. 2641--2649.

\bibitem{resnet}
K.~He, X.~Zhang, S.~Ren, and J.~Sun, ``Deep residual learning for image recognition,'' in \emph{2016 IEEE Conference on Computer Vision and Pattern Recognition (CVPR)}, 2016, pp. 770--778.

\bibitem{swin}
Z.~Liu, Y.~Lin, Y.~Cao, H.~Hu, Y.~Wei, Z.~Zhang, S.~Lin, and B.~Guo, ``Swin transformer: Hierarchical vision transformer using shifted windows,'' in \emph{Proceedings of the IEEE/CVF International Conference on Computer Vision (ICCV)}, 2021.

\bibitem{fixedAP}
A.~Dave, P.~Doll{\'a}r, D.~Ramanan, A.~Kirillov, and R.~Girshick, ``Evaluating large-vocabulary object detectors: The devil is in the details,'' \emph{arXiv preprint arXiv:2102.01066}, 2021.

\end{thebibliography}

\end{document}